\documentclass[10pt,twocolumn,letterpaper]{article}

\usepackage{cvpr} 

\usepackage{siunitx}
\usepackage{makecell}

\usepackage{epsfig}
\usepackage{amsmath}
\usepackage{amssymb}
\usepackage{booktabs}
\usepackage{multirow}
\usepackage[ruled,vlined]{algorithm2e}
\usepackage{pifont}
\usepackage[dvipsnames, svgnames, x11names]{xcolor}         

\usepackage{colortbl}       
\usepackage{pifont}
\usepackage{graphicx} 

\usepackage{algorithmic}
\usepackage[dvipsnames, svgnames, x11names]{xcolor}         
\newcommand{\website}{\url{https://humanoid-exo.github.io/}}

\definecolor{cvprblue}{rgb}{0.21,0.49,0.74}
\usepackage[pagebackref,breaklinks,colorlinks,allcolors=cvprblue]{hyperref}

\def\paperID{} 
\def\confName{CVPR}
\def\confYear{2025}

\title{HumanoidExo: Scalable Whole-Body Humanoid  Manipulation\\ via Wearable Exoskeleton}

\author{
\text{Rui Zhong$^{1,}$}\thanks{: Co-first author. $\dagger$: Corresponding Author. This work was done during Rui Zhong’s internship at Midea Group.}
\text{, Yizhe Sun$^{2}$}, 
\text{Junjie Wen$^{2}$},
\text{Jinming Li$^{2}$},
\text{Chuang Cheng$^{1,\dagger}$},
\\
\text{Wei Dai$^{1}$},
\text{Zhiwen Zeng$^{1}$},
\text{Huimin Lu$^{1,\dagger}$},
\text{Yichen Zhu$^{2,*,\dagger}$}, 
\text{Yi Xu$^{2}$}
\\
\tt\small $\textsuperscript{1}\text{National University of Defense Technology}$, $\textsuperscript{2}\text{Midea Group}$
\\
\hspace{0cm}\large\website
\vspace{-0.5cm}
}

\begin{document}

\makeatletter
\let\@oldmaketitle\@maketitle
\renewcommand{\@maketitle}{\@oldmaketitle
    \begin{center}
        \captionsetup{type=figure}
        \centering
        \includegraphics[width=0.98\textwidth]{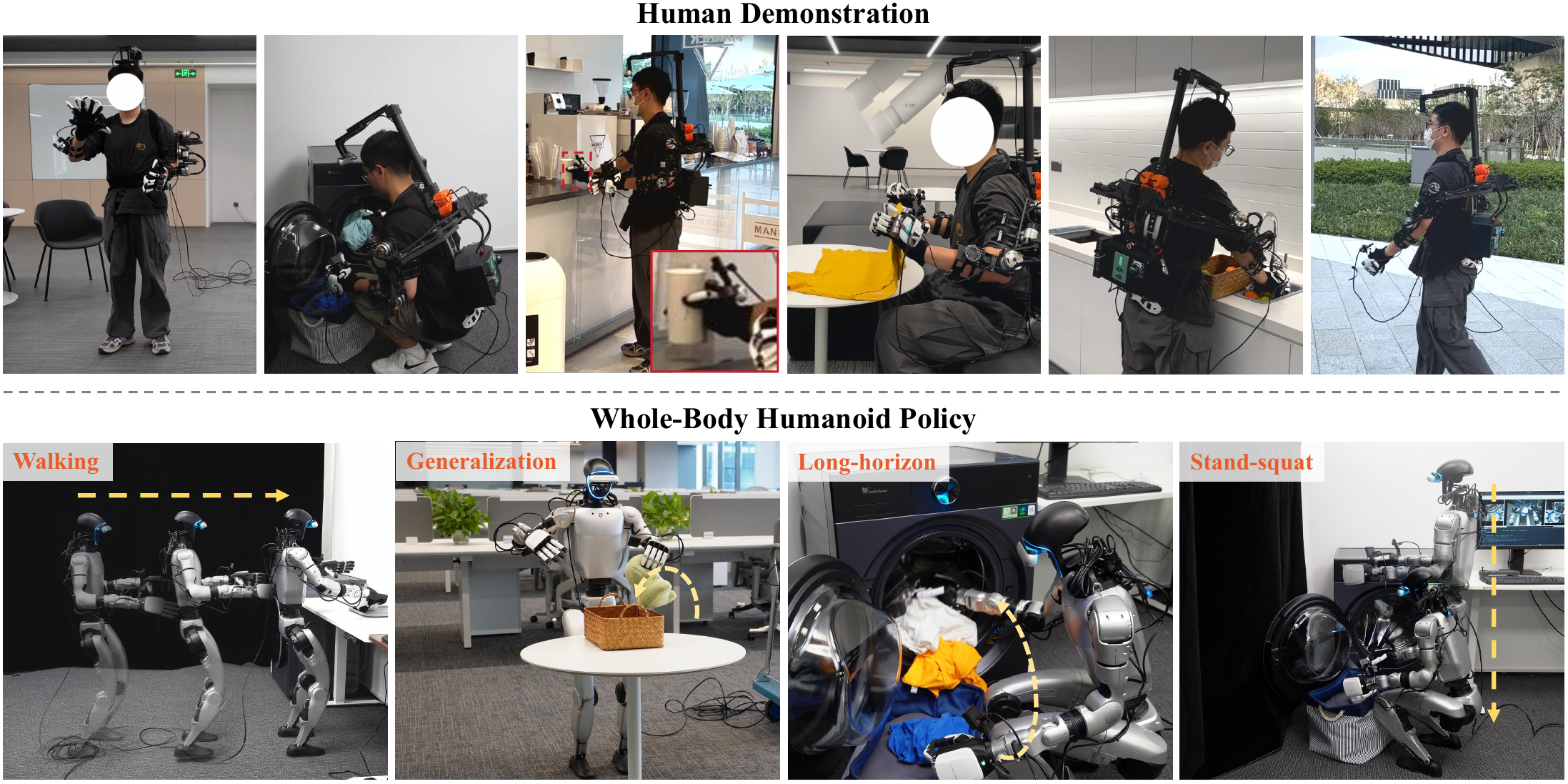}
        \caption{\textbf{HumanoidExo}, a wearable exoskeleton system that transfers human motion to \textbf{whole-body} humanoid data. HumanoidExo provides an efficient solution that bridges the embodiment gap between humans and robots, facilitating the collection of diverse datasets. We tested it on three real-world tasks, and the results show that it helps humanoid robots generalize to new environments, learn complex tasks from limited data, and acquire new skills like walking.}
    \end{center}
}
\makeatother

\maketitle

\begin{abstract}
A significant bottleneck in humanoid policy learning is the acquisition of large-scale, diverse datasets, as collecting reliable real-world data remains both difficult and cost-prohibitive. To address this limitation, we introduce \textit{HumanoidExo}, a novel system that \textit{transfers human motion to whole-body humanoid data}. HumanoidExo offers a high-efficiency solution that minimizes the embodiment gap between the human demonstrator and the robot, thereby tackling the scarcity of whole-body humanoid data. By facilitating the collection of more voluminous and diverse datasets, our approach significantly enhances the performance of humanoid robots in dynamic, real-world scenarios. We evaluated our method across three challenging real-world tasks: table-top manipulation, manipulation integrated with stand-squat motions, and whole-body manipulation. Our results empirically demonstrate that HumanoidExo is a crucial addition to real-robot data, as it enables the humanoid policy to generalize to novel environments, learn complex whole-body control from only five real-robot demonstrations, and even acquire new skills (i.e., walking) solely from HumanoidExo data. 
\end{abstract}

\section{Introduction}
Humanoid policy learning is a rapidly advancing field, now spanning locomotion, manipulation, and language-conditioned tasking. This progress is largely propelled by foundational model initiatives for general-purpose humanoids, such as Nvidia's GR00T\cite{nvidia2025gr00tn1} and Figure AI. To mitigate the high cost of real-robot demonstrations, researchers have introduced several data-efficient pipelines. These include sim-to-real transfers, learning from web-scale human videos (e.g., EgoMimic\cite{Kareer2025EgoMimic}, Vid2Robot\cite{jain2024vid2robot}), and the development of diverse teleoperation systems for more effective data collection.

Despite these efforts, scaling humanoid data collection remains a significant challenge for two primary reasons. First, simulation and human video data both suffer from severe embodiment gaps. Simulated robot dynamics inevitably mismatch their real-world counterparts, while the morphological and kinematic differences between humans and robots make direct video-to-policy transfer notoriously difficult. Second, direct teleoperation is difficult to scale. This approach typically requires a one-to-one, human-to-robot setup, which is expensive and resource-intensive. Furthermore, the process is physically and mentally demanding, leading to operator fatigue that limits the duration of collection sessions and requires highly skilled personnel. This reliance on expert operators and the inherent risk of damaging costly hardware during live sessions create a major bottleneck for generating large-scale datasets.

In this work, we introduce HumanoidExo, an integrated system that advances both data collection and policy learning for humanoid robotics. We utilize a custom-designed, lightweight, and flexible wearable exoskeleton\cite{zhong2025nuexo} to capture human motion without impeding the operator's natural movements. This design enables the comfortable performance of diverse daily tasks, while our system records and translates the operator's actions into structured data for robot learning. To capture comprehensive, whole-body motion, a back-mounted LiDAR sensor tracks the operator's torso, providing a 6D pose to record base movements such as walking, squatting, and bending. By fusing data from the exoskeleton and LiDAR, our system generates kinematically feasible, whole-body trajectories ready for large-scale policy learning.

To leverage this data, we present \textbf{H}umanoid\textbf{E}xo-VLA (\textbf{HE-VLA} in short), a refined, vision-language-action model for whole-body humanoid policy learning from exoskeleton data. This hybrid approach uses imitation learning as a foundation and incorporates reinforcement learning to ensure the robot maintains balance and stability during movement and manipulation. The synergy between our hardware and software enables stable, efficient policy learning from exoskeleton data, leading to policies that are directly deployable on a physical humanoid robot.

To validate the effectiveness of our method, we conduct a comprehensive study across three challenging real-world tasks. These tasks include table-top manipulation, dexterous manipulation involving stand-squat motions, and whole-body manipulation that requires walking to a table. Our experimental results highlight the critical role of HumanoidExo data in enhancing policy performance: 1) \textbf{Generalization}: It enables the learned policy to generalize effectively to novel scenes and environments. 2) \textbf{Data Efficiency}: It allows an end-to-end model to learn complex tasks with as few as five real-robot demonstrations. 3) \textbf{Skill Acquisition}: It empowers the humanoid robot to acquire entirely new skills (e.g., walking) using only data from the exoskeleton, without any real-robot demonstrations. We believe that HumanoidExo represents a significant step toward achieving scalable whole-body humanoid policy learning.

\section{Related Work}
\subsection{Data Collection with In-The-Wild Equipment}
To enable scalable robot data collection, several in-the-wild systems have recently been proposed to make the process more affordable and accessible~\cite{bahl2022human,ye2025video2policy,liu2024fastumi,Kareer2025EgoMimic,fang2025dexop,liu2025immimic}. These systems have shown great promise in specific domains. For instance, Human Policy~\cite{qiu2025humanoidpolicy} focuses on transferring human action primitives to bridge embodiment differences. DexCap~\cite{wang2024dexcap} uses a wearable glove to capture precise wrist and fingertip poses for dexterous tasks. AirExo\cite{fang2024airexo,fang2025airexo2} leverages low-cost hardware with direct kinematic mapping for arm manipulation. The Universal Manipulation Interface (UMI)~\cite{chi2024umi} introduced a simple handheld controller for collecting bimanual data at scale, which DexUMI~\cite{xu2025dexumi} later extended to dexterous hands. However, these pioneering works share two common limitations. First, they are primarily designed for conventional robotic arms. Second, their focus is typically restricted to upper-body manipulation. In contrast, our work, HumanoidExo, addresses these gaps directly. To the best of our knowledge, this is the first system designed for in-the-wild, whole-body policy learning on a humanoid robot. We introduce multiple techniques to bridge the embodiment gap between humans and humanoids, effectively translating natural human motion into executable robot policies. For the first time, we demonstrate that this paradigm can successfully scale up the acquisition of whole-body training data, paving the way for more capable and generalist humanoid robots.

\subsection{Humanoid Whole-Body Manipulation}
Reinforcement learning algorithms for whole-body locomotion of humanoid robots have been extensively studied\cite{huang2025learnstandingup,Rado2024Realworld,zhang2025falcon,zhang2024Whole-bodyHumanoid}, and whole-body teleoperation of humanoid robots can be achieved through various methods\cite{ze2025twist,sun2025ulc,li2025clone,ben2024homie,cheng2024opentelevision,fu2024humanplus,jiang2025behaviorrobotsuite,li2025amo}. In robotic manipulation, imitation learning~\cite{act, chi2023diffusion_policy,chi2024umi,he2024omnih2o,zhang2025doglove,Zhu2025scaledp} with rich visual representations, especially from robot foundation models like Vision-Language-Action (VLA) models~\cite{wen2025dexvla, black2024pi0, intelligence2025pi05,GeminiRobotics, wen2024tinyvla, zhou2025chatvla, zhou2025chatvla2,yang2025egovla,wang2025trackvla,liu2025hybridvla,li2025pointvla,deng2025graspvla,li2025controlvla}, has become a central paradigm. This line of work typically focuses on humanoid manipulation~\cite{ding2025humanoidvla}; for instance, iDP3~\cite{ze20243d} trains egocentric 3D diffusion policies, and Dexmimicgen~\cite{jiang2025dexmimicgen} learns bimanual skills via a sim-to-real strategy. These methods mainly rely on teleoperated data collected directly from the physical humanoid robot. In contrast, our approach, HumanoidExo, focuses on learning humanoid control through a novel mixture of limited real robot data and a larger dataset collected from a human operator wearing an exoskeleton. More importantly, our method is not confined to manipulation but extends to complex whole-body control tasks such as standing, squatting, and walking.

\begin{figure*}[t]
    \centering
    \includegraphics[width=1\linewidth]{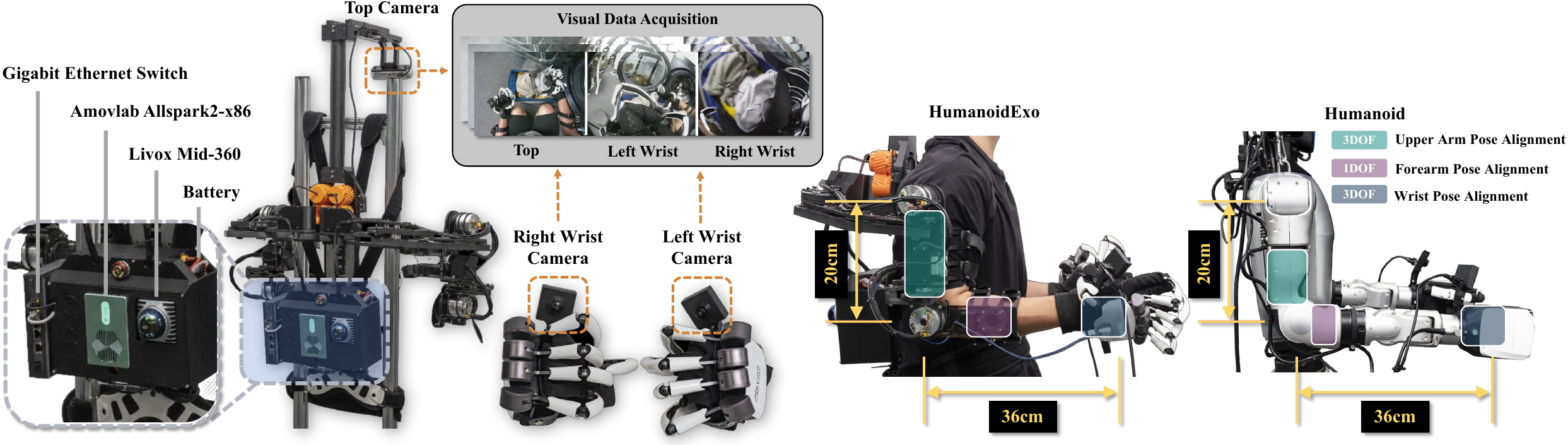}
    \caption{\textbf{Hardware overview for HumaniodExo}. We integrated a Mid-360 LiDAR for acquiring exoskeleton motion odometry. For visual information acquisition, we added two wrist cameras to capture new operational perspectives and enrich environmental perception. These cameras, installed on the Dexmo force-feedback gloves, were mounted at angles identical to those of the robot’s cameras. Since the HumanoidExo system adopts a joint space control method with angle remapping, we redesigned the exoskeleton’s key parameters to match the arm length of the Unitree G1 robot. In addition, we recruited data collectors with upper-body dimensions similar to the G1’s anthropometric parameters for data collection and teleoperation, minimizing end-effector errors arising from dimensional mismatches.}
    \label{fig:hardware}
\end{figure*}

\section{Methodology}
This section provides a comprehensive technical overview of our proposed HumanoidExo system. We begin by elucidating the foundational concept of our approach, which is designed to bridge the significant embodiment gap between a human demonstrator and a humanoid robot. Then, we detail the specific implementation for data acquisition, breaking it down into our methodologies for capturing and refining upper-body motion and for tracking lower-body dynamics. Subsequently, we describe the motion retargeting pipeline that integrates these data streams and adjusts the complete whole-body trajectories for effective robot data. To conclude, we specify the hardware configuration of the target humanoid platform used in our experiments.

\subsection{Bridging the Embodiment Gap Between Human and Humanoid}
One shortcoming that preventing scaling humanoid data with exoskeleton or universal interface control is let in the control mapping methods. Specifically, robot arm control typically involves two primary methods: Cartesian space control (end-effector control) and joint space control. For devices such as the Meta Quest or Apple Vision Pro, which are capable of capturing the end-effector pose, Cartesian control is often an appropriate choice. However, in scenarios where precise arm joint angles are required—such as when placing clothes in a washing machine—solely tracking the end-effector pose without considering the other arm joints can easily result in collisions with the environment. Moreover, since a typical humanoid robot arm has seven degrees of freedom (DoF), its inverse kinematics has multiple solutions, leading to a null space. End-effector control in such cases necessitates additional constraints, making real-time and accurate joint angle computation particularly challenging in small, confined spaces. Therefore, it is more practical to directly map the joint space, aligning human joint movements with those of the robot.

\subsection{Upper-Body Alignment between HumaniodExo and Robot}
HumanoidExo is specifically designed to read all seven joints of the human arm. The rotational axes of its exoskeleton arm are precisely aligned with the corresponding axes of the human joints, making the exoskeleton isomorphic to the human arm. Additionally, we have introduced two extra DoF at the Glenohumeral (GH) joint, enhancing the system's ergonomics and making it more suitable for daily wear~\cite{kim2017ijrr}. Due to these extra DoF and the incorporation of a new link system, HumanoidExo’s joint configuration differs significantly from that of a typical humanoid robot arm. The motion retargeting algorithm used for this process is outlined below.

\textbf{Upper Arm Pose Alignment:} The exoskeleton’s base is used as the reference coordinate system, fixed to the wearer’s torso. To align the upper arm pose, we first extract all rotational joints leading to the upper arm attachment point. This includes both active joints driven by motors and passive joints within the linkages and timing belts. By analyzing the relative positions of these joints, we construct a Denavit-Hartenberg (DH) parameter table\cite{zhong2025nuexo}. Using forward kinematics, we then compute the relative pose between the upper arm attachment point and the base, expressed as a quaternion ~${\boldsymbol{q}}^{upper\, arm}$. Subsequently, an iterative inverse kinematics method is employed to map the upper arm’s pose onto the first three joints of the robot’s arm, thereby achieving upper arm pose alignment.

\textbf{Forearm Pose Alignment:} The elbow is the only moving joint between the human upper arm and forearm. HumanoidExo is equipped with a motor at the elbow, enabling direct joint space control. This motor allows us to map the human elbow's bending angle directly to the robot’s forearm, achieving forearm pose alignment.

\textbf{Wrist Pose Alignment:} The human wrist can be treated as a spherical joint\cite{zimmermann2023tro}. To simplify the structural design and reduce the wrist's weight while still enabling the exoskeleton to capture wrist pose, we utilize Inertial Measurement Units (IMUs) as the joint data source. To capture the wrist’s three rotational DoF relative to the forearm, we place one IMU on the forearm and another on the Dexmo force feedback glove. First, we record the quaternions of the wrist IMU,~${\boldsymbol{q}}^{w}$, and the forearm IMU,~${\boldsymbol{q}}^{f}$. 
Upon initializing the exoskeleton, the operator returns the wrist to the home position. Now record both IMU readings at this time ${\boldsymbol{q}}^{w_{0}}$ and ${\boldsymbol{q}}^{f_{0}}$, The rotation quaternion $\boldsymbol{q}^{wrist}$ representing the wrist's orientation relative to the forearm at any given moment:
\[
{\boldsymbol{q}^{wrist}} = {{\boldsymbol{q}}^{t}}{\boldsymbol{q}}^{t_0} = {{\boldsymbol{q}}^{f}}^{*}{{\boldsymbol{q}}^{w}}{{\boldsymbol{q}}^{w_{0}}}^{*}{{\boldsymbol{q}}^{f_{0}}}
\]
where~${{\boldsymbol{q}}}^{*}$~represents the conjugate of the quaternion. This quaternion serves as the target pose for the wrist, and an iterative inverse kinematics method is employed once more to map this pose to the final three joints of the robot’s arm, achieving wrist pose alignment.

\begin{figure*}[htb]
  \centering
  \includegraphics[width=1\linewidth]{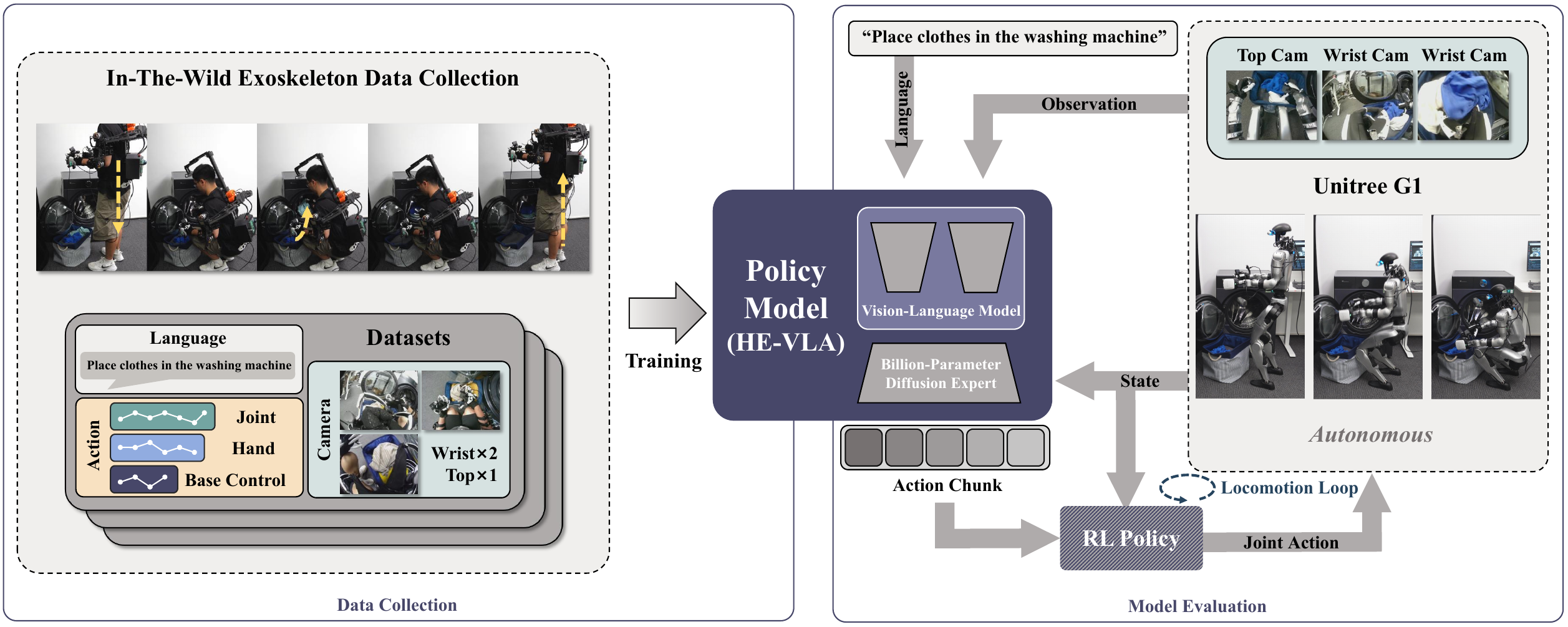}
  \caption{\textbf{The overview of the HE-VLA}. The left side of the figure illustrates our data collection process and the composition of the dataset, with the primary source of training data being in-the-wild data gathered by the HumanoidExo independently of the controlled robot. The right side of the figure presents the model deployment and inference pipeline. We employ two systems to control the robot: a Vision-Language-Action (VLA) model and a Reinforcement Learning (RL) model. The VLA model generates high-level control commands and transmits them to the lower-level RL model. The RL model is then responsible for maintaining balance control and executing the joint movement commands.}
  \label{fig:system}
\end{figure*}

\begin{figure}[t]
  \centering
  \includegraphics[width=1\linewidth]{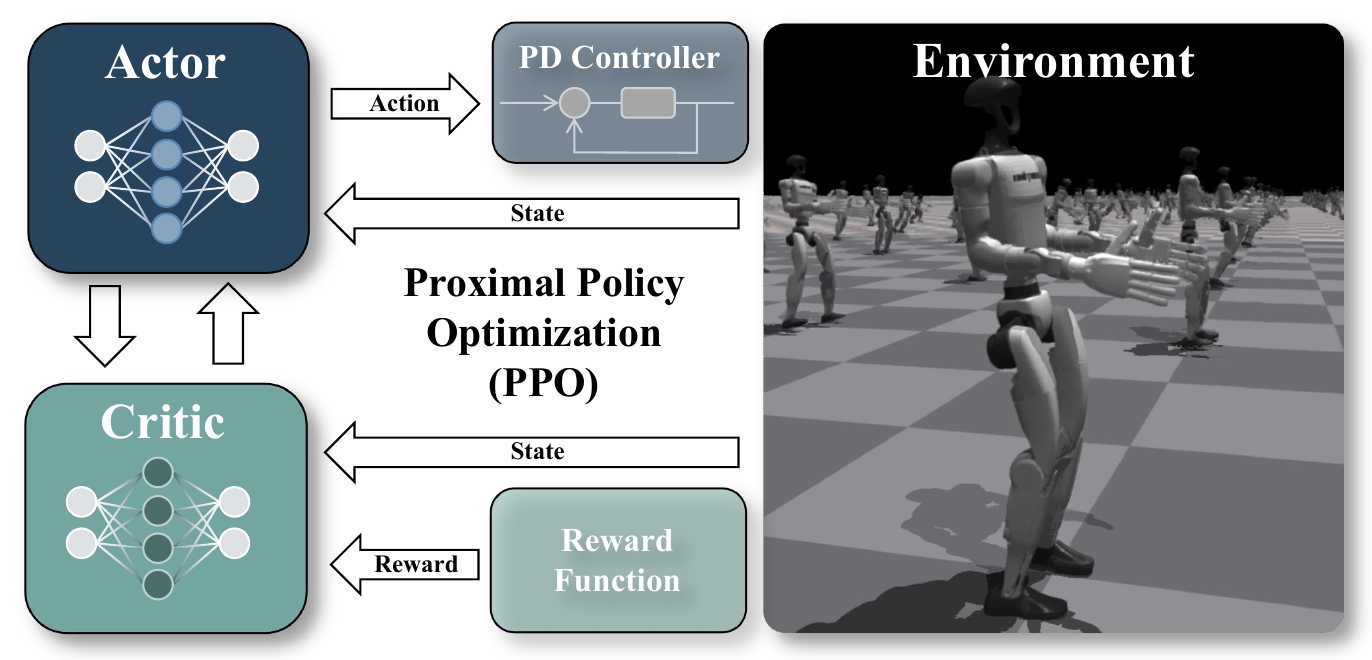}
  \caption{\textbf{The actor-critic reinforcement learning in HE-VLA}. This module works in conjunction with the primary VLA model, guaranteeing the humanoid can reliably stand, squat, and walk during policy inference.}
  \label{fig:RL}
\end{figure}

\subsection{Lower-Body Alignment between HumaniodExo and Robot}
For humans, lower-limb movements such as walking, squatting, and standing are fundamental to daily activities, as they directly determine mobility, flexibility, and the ability to interact with complex environments. Motivated by this, our system adopts base translational velocity, base rotational velocity, and height as the command for the RL-based balance control algorithm. These variables capture the essential DoF of human lower-body motion, enabling intuitive and effective control while also providing a foundation for high-level task planning.

Beyond supporting whole-body teleoperation of humanoid robots, our HumanoidExo system is designed with an independent data collection capability. Specifically, it can perceive and record its own motion state in real time—including displacement, orientation, and variations in body height—without depending on external operator input.

To realize this, we integrated a Mid-360 LiDAR on the back of the exoskeleton and developed a LiDAR-odometry module based on FAST-LIO\cite{xu2021fastlio} for real-time estimation of the system’s spatial position and orientation. Compared with approaches relying on vision or external localization beacons, LiDAR odometry provides superior environmental adaptability and robustness to lighting conditions. It maintains high localization accuracy even under drastic illumination changes or in texture-sparse environments. Moreover, it eliminates the need for external infrastructure, thereby enabling truly independent operation and reliable data collection in open, unstructured, or entirely unknown in-the-wild scenarios.

The specific mounting position and configuration of the LiDAR are illustrated in Fig. \ref{fig:hardware} (left). The placement was carefully designed to balance field of view, resilience to interference, and overall motion stability of the system.

\subsection{HE-VLA: A Whole-Body Humanoid Policy Learning Method}
This section introduces our method for learning a whole-body humanoid policy from our collected exoskeleton data. Our approach, namely \textbf{H}umanoid\textbf{E}xo-VLA (\textbf{HE-VLA} in short), consists of two key components: a pre-trained Vision-Language-Action (VLA) model that learns foundational whole-body motion control, and a reinforcement learning method that ensures robust whole-body balance.

\textbf{Vision-Language-Action Model.} Given a set of expert demonstrations that contain complex humanoid skill trajectories, we want to learn a visuomotor policy $\pi : \mathcal{O} \mapsto \mathcal{A}$ that maps the visual observations $o \in \mathcal{O}$ to actions $a \in \mathcal{A}$, such that our robots not only \textit{reproduce} the skill but also \textit{generalize} beyond the training data.
To tackle the challenge of manipulating complex humanoid skills, we leverage DexVLA\cite{wen2025dexvla}, a pre-trained vision-language-action model, for the tasks described in our experiments. Notably, the pre-training dataset for this foundation model does not include data from the humanoid robot used in our experiments, the Unitree G1. By leveraging a pre-trained robot foundation model, we ensure its parameters are effectively warmed up, facilitating an easier acquisition of complex real-world skills. However, we observe that finetuning the model alone does not guarantee stable whole-body control. In fact, the fine-tuned model often causes the robot to fall, failing to maintain whole-body stability due to instability induced by dynamic upper-body motions. Therefore, we introduce an improvement to address this critical problem.

\textbf{Reinforcement Learning for Motion Balance.} Relying solely on imitation learning to directly output joint positions for whole-body control introduces significant stability risks. Minor deviations from the learned trajectories can result in falls, posing a threat of catastrophic damage to the robot and its environment. To overcome this limitation, we leverage reinforcement learning to train a robust whole-body loco-manipulation controller. This controller is responsible for maintaining dynamic balance while executing commands for base speed, yaw rate, and a target torso height.

Specifically, at each step, the policy receives
$
\mathbf{o}_t=\big[c_t,\ \boldsymbol{\omega}_t^b,\ \mathbf{g}_t^b,\ \mathbf{q}_t,\ \dot{\mathbf{q}}_t,\ \mathbf{a}_{t-1}\big]$, and $c_t=\big[v_x^\ast,\ \omega_z^\ast,\ h^\ast\big],
$
where $\boldsymbol{\omega}_t^b$ and $\mathbf{g}_t^b$ are base angular velocity and gravity in the torso frame, and $(\mathbf{q}_t,\dot{\mathbf{q}}_t)$ are joint states. The action $\mathbf{a}_t$ specifies desired lower‑body joint targets $\mathbf{q}^{\mathrm{des}}=\mathbf{q}^{0}+\mathbf{a}_t$. Joint torques are applied by a PD law
\[
\tau_i=k_{p,i}\!\left(q^{\mathrm{des}}_i-q_i\right)-k_{d,i}\dot q_i,
\]
driving locomotion while the upper‑body joints follow the operator directly.

To achieve stable walking with dynamic upper-body movement, we use a curriculum that scales the admissible joint range by a ratio $r \in [0,1]$. This ratio increases as the agent succeeds, smoothly expanding motion from static to fully expressive. Commanded squats are realized through a reward function that tracks a target base height and shapes knee flexion accordingly. We train this by dedicating one-third of our parallel environments to squatting and the rest to walking, periodically switching their roles to learn seamless transitions. Each transition is mirrored across the robot’s $x$–$z$ plane (swap left/right states and actions and flip the yaw command), and both original and mirrored samples are stored. Auxiliary actor/critic symmetry losses encourage consistent predictions on mirrored pairs, improving sample efficiency and reducing unintended left–right bias. As a result, the learned policy walks, turns, and squats to commanded heights while remaining stable under continuously changing operator-driven upper-body poses.

\subsection{Robot Configuration}
\textbf{Camera View.} The visual data acquisition system is composed of a primary head-mounted camera (Realsense D455) and two supplementary wrist-mounted fisheye cameras, which offer expanded perspectives during manipulation. To mitigate the visual embodiment gap, we standardized the hardware by using the same camera models on both the exoskeleton and the robot. Furthermore, the field of view (FoV) of the wrist cameras on the exoskeleton is carefully calibrated to match that of the robot's cameras.

We identified two primary challenges related to the head camera. First, the Unitree G1 robot features a non-actuated head, making it impossible to translate the human operator's head movements into robot camera motion. Second, a perfect positional correspondence between the exoskeleton's and the robot's head cameras is not achievable. Nevertheless, our empirical results demonstrate that these discrepancies do not hinder policy learning; the model proves capable of extracting salient visual features even when trained with the non-stationary head camera view from the human demonstrator.

\textbf{Robot Setup.} Our experimental platform is the Unitree G1, a humanoid robot with a height of 1.3m and 29 DoF in its body. The robot is equipped with two Inspire-Hand end-effectors. Each hand is an underactuated gripper with 12 DoF (six active and six passive). The thumb contains two active and two passive DoFs, while each of the remaining fingers has one active and one passive DoF. As a result, the entire robotic system has a total of 41 active degrees of freedom. Our policy model outputs target joint angles to control the robot's entire body.

\section{Experiment}
\begin{figure*}[htb]
  \centering
  \includegraphics[width=1\linewidth]{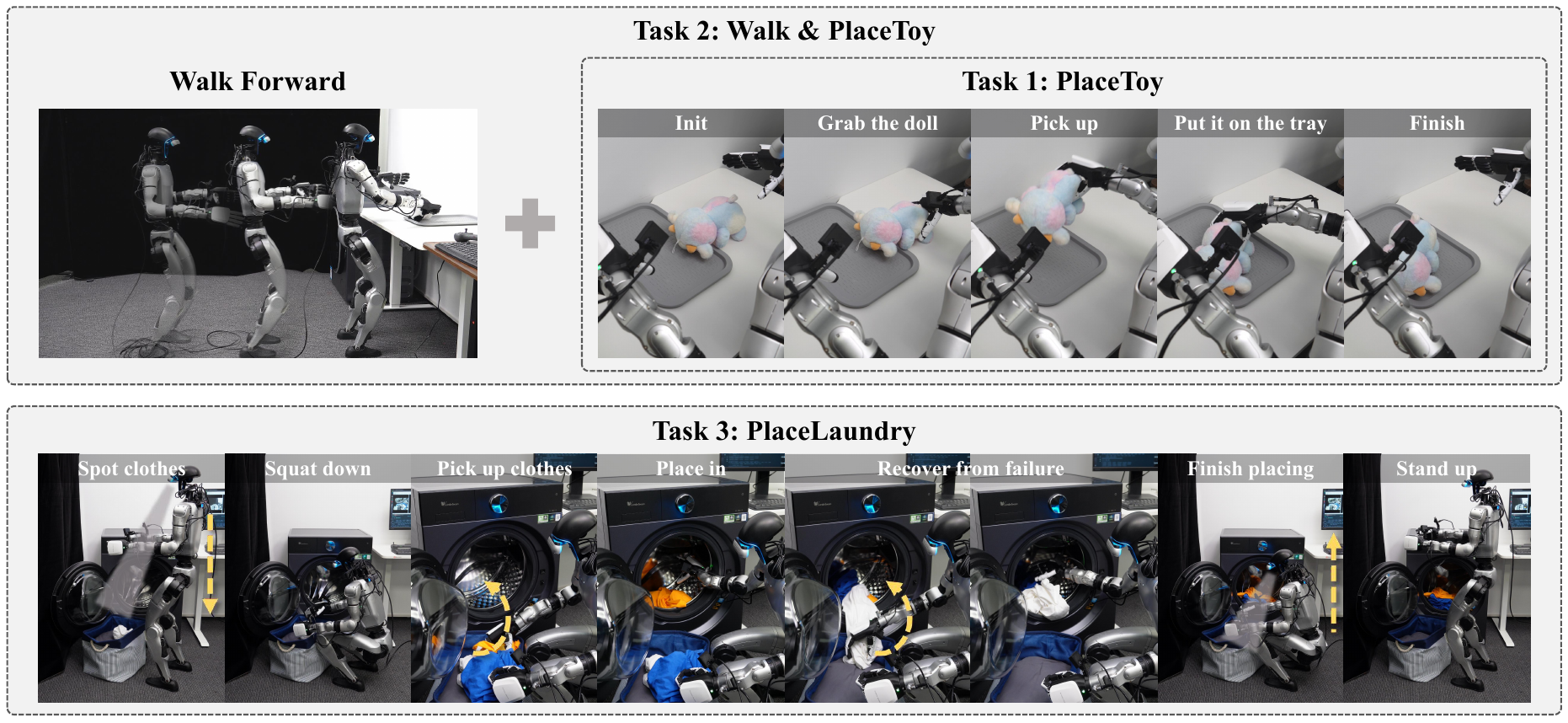}
  \caption{\textbf{Examples for PlaceToy (Task 1), Walk \& PlaceToy (Task 2), and PlaceLaundry (Task 3).} We designed three tasks to showcase the effectiveness of HumanoidExo in robot skill learning: Task 1 tests dexterity, Task 2 combines locomotion and manipulation, with the mobile-manipulation data entirely collected by HumanoidExo, and Task 3 involves whole-body manipulation of the humanoid robot.}
  \label{fig:all_exp}
\end{figure*}

In the experiments section, we aim to answer the following questions to demonstrate the effectiveness of HumanoidExo:

\begin{itemize}
\item How does the HumanoidExo system compare to state-of-the-art policy learning methods?

\item Does training with HumanoidExo data improve the policy's robustness to environmental variations and physical disturbances?

\item Can HumanoidExo enable the learning of novel and complex skills for humanoid robots?

\item To what extent does exoskeleton data facilitate learning skills that involve mobility and locomotion?

\end{itemize}

\subsection{Task Descriptions \& Implmeentation Details}
As shown in Fig. \ref{fig:all_exp}, HumanoidExo performs on challenging whole-body humanoid manipulation tasks. We conducted experiments on three tasks that represent different categories of humanoid skills.
\begin{itemize}
    \item \textbf{PlaceToy}: This is a tabletop manipulation task. The robot is required to pick up a toy, whose position is randomized on its left or right side, and place it into a tray at the center. This task tests dexterity without requiring locomotion.

    \item \textbf{Walk \& PlaceToy}: This task combines locomotion with manipulation. The humanoid is required to walk to a table, stop, and then place a toy into a tray at the center.
        
    \item \textbf{PlaceLaundry}: This is a whole-body manipulation task. The humanoid must squat down, pick up an article of clothing from a basket, place it into a washing machine on its right, and then return to a standing position. The primary challenge is maintaining balance during the dynamic squatting and standing motions.

\end{itemize}
We give examples for each task in Figure~\ref{fig:all_exp}. For all experiments, we fine-tune the model for 50k iterations with a batch size of 128. We use a cosine learning rate scheduler with an initial learning rate of 2e-5. All models and experiments use the same set of hyperparameters to ensure fair comparison.

\subsection{Generalizable Manipulation with HumanoidExo}
\label{sec:tabletop_exp}
The experiments in this section aim to measure the efficiency of HumaniodExo data and the generalizability that such data bring. 

\begin{figure}[t]
  \centering
  \includegraphics[width=1\linewidth]{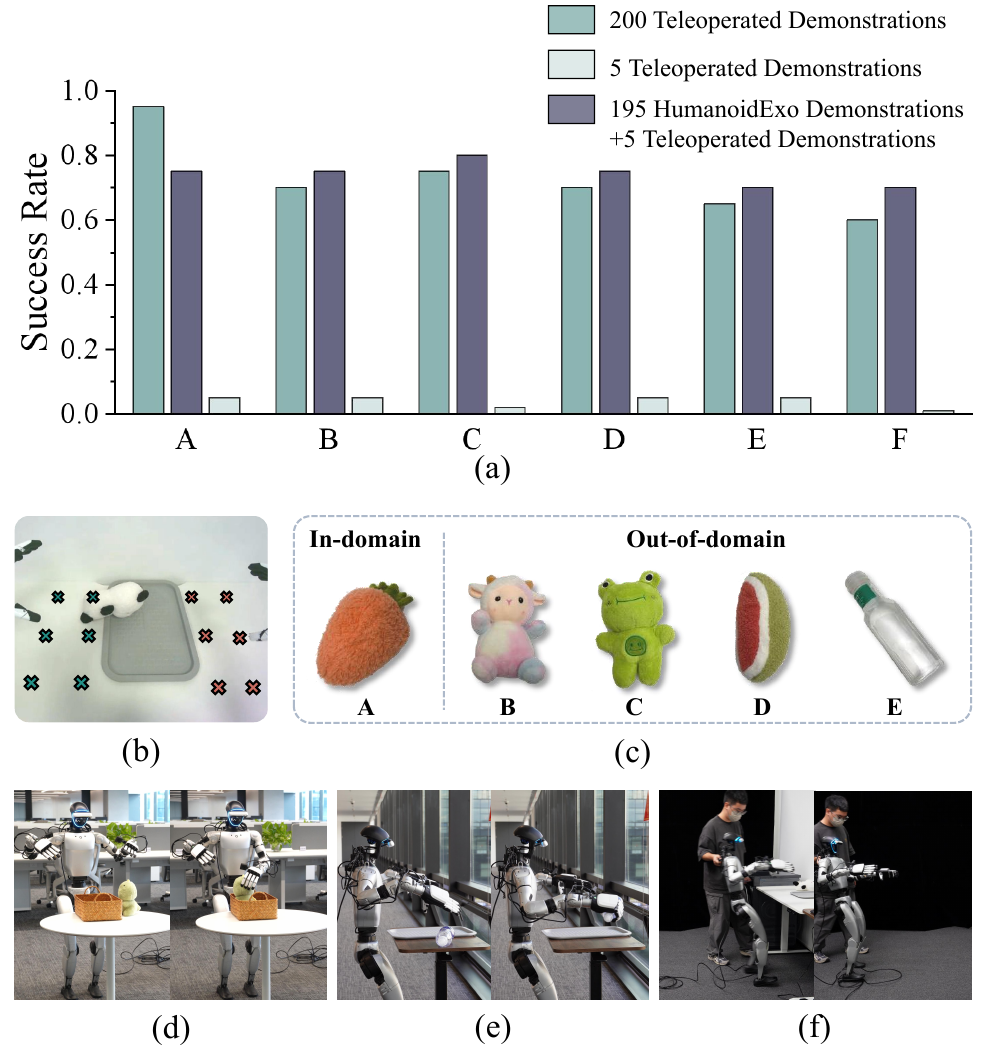}
  \caption{\textbf{Model Generalization.}
    \textbf{(a) Model success rates.} Labels A-E correspond to the robot's success rate for grasping items A-E shown in (c), and the number of trials for each experiment is 60.
    \textbf{(b) Item placement locations for model testing.}
    \textbf{(c) Training datasets.} Out-of-domain data represents items that did not appear in the Teleoperated Demonstrations but were present in the HumanoidExo Demonstrations.
    \textbf{(d)\&(e) Tasks in new environment.} The robot's success rate for completing the task is represented by the label F in (a).
    \textbf{(f) Robustness to disturbance.} The robot could autonomously walk back to the table and resume the tabletop task (Task 2) after being forcibly moved away.}
  \label{fig:generalization_exp}
\end{figure}

Fig.~\ref{fig:generalization_exp} summarizes the average success rates for the tabletop manipulation task across three experimental setups: 1) training with 200 real teleoperated demonstrations; 2) mixed training with 5 teleoperated and 195 HumanoidExo demonstrations; 3) training with only 5 teleoperated demonstrations. This experiment aims to evaluate the model's performance under an extreme data-scarce scenario, where real data constitutes just 2.5\% of the mixed dataset.

The results reveal two key findings. First, HumanoidExo data provides a significant performance boost when augmenting a small set of real demonstrations. For instance, while training with only five teleoperated demonstrations yields a mere 5\% success rate, adding 195 HumanoidExo demonstrations boosts the success rate dramatically to about 80\%. Second, for this in-domain task, the performance achieved with the mixed dataset suggests that HumanoidExo data has a utility comparable to that of real-robot data. This is particularly significant because collecting HumanoidExo data is substantially more cost-effective and scalable than robot teleoperation, highlighting the practical value and efficiency of our method.

\textbf{Object \& Environment Generalization.}
A key metric for evaluating a policy is its ability to generalize to novel objects and unseen environments. To assess this, we conduct two generalization tests: one involving four objects not present in the training data, and another in a completely new environment with different objects (a novel toy and tray).

Across all generalization settings, we observe that a mixed training approach—combining teleoperated data with HumanoidExo data—yields a higher average success rate than training solely on teleoperated data. These findings demonstrate the effectiveness of our proposed system in improving policy generalization.

\subsection{Exoskeleton Data Brings New Skill}
Given that HumanoidExo data can serve as an effective substitute for teleoperated data, a natural question arises: can it also be used to teach the robot entirely new skills?

To investigate this, we designed an experiment that builds upon the previous tabletop manipulation setup. We collected 195 new HumanoidExo demonstrations of a compound task: walking to the table, stopping, and then executing the 'place toy' action. The policy was then trained on a mixed dataset containing these 195 new HumanoidExo demonstrations and the same five teleoperated demonstrations from the previous experiment. Crucially, these five teleoperated demonstrations only contain the stationary manipulation portion of the task; they include no walking. Therefore, any walking ability exhibited by the final policy must be learned exclusively from the HumanoidExo data. The task is shown in Figure~\ref{fig:all_exp}.

For this combined task, the policy achieved a 100\% success rate on the walking portion in every trial. The robot consistently navigated to the table and stopped in the correct position before initiating the manipulation phase. Consequently, the overall task success rate is identical to that of the stationary pick-and-place experiment. This result demonstrates that our method successfully empowered the robot with the new skill of walking without degrading its previously learned manipulation capabilities. This promising phenomenon motivated us to conduct further tests on the robustness and generalization of this newly acquired skill, which was learned exclusively from HumanoidExo data.

\textbf{Robustness to Disturbance.} An interesting observation from our experiments is the trained policy's remarkable robustness to physical disturbances. As demonstrated in Figure~\ref{fig:generalization_exp}(f), when we manually drag the Unitree G1 away from its workspace, the policy consistently recovers by walking the robot back to the table.

Notably, the model can recover even when displaced to distances far exceeding any seen in the HumanoidExo training data. It successfully navigates back to its station in front of the table before proceeding to complete the task. This behavior suggests that training with HumanoidExo data instills the policy with strong generalization capabilities against significant real-world perturbations.

\textbf{Generalization to New Environment.} To evaluate generalization, we tested the policy in a new environment, as demonstrated in the supplementary video. The robot was then tasked with performing the same "walk and place" routine. While we observed a slight decrease in the success rate for the manipulation phase, the walking skill transferred perfectly to this unseen setting. This result demonstrates the strong generalization ability of the locomotion capabilities learned through our approach, even when the visual context changes significantly.

\subsection{HumanoidExo for Whole-Body Deterxous Manipulation}
We conducted further experiments to test the whole-body dexterous manipulation capabilities of our humanoid robot. Specifically, we chose a "Place Laundry" task, where the robot is required to squat, grasp clothes from a basket, and place them into a washing machine on its right. The robot repeats this process until the basket is empty, then stands up to signal task completion. This task presents several challenges: the clothes are deformable objects that are difficult for dexterous hands to grasp and place entirely inside the machine, which requires the model to exhibit recovery behaviors. Furthermore, the model must use robust visual observation while maintaining whole-body balance to avoid falling during the upper-body task execution. 

Following the methodology of the previous section, we trained the HE-VLA model in three different configurations: one with 200 teleoperated demonstrations, a second with only 5 teleoperated demonstrations and 195 HumanoidExo data points, and a third with only 5 teleoperated demonstrations. The results, presented in Table~\ref{tbl:PlaceLaundry}, show that the model trained on 200 teleoperated demonstrations performs similarly to our model, which used 195 HumanoidExo data points. These results demonstrate that even for complex whole-body dexterous manipulation, our method can be a strong replacement for teleoperated data and can even be more effective for task generalization.

\begin{table}[t]
\centering
\caption{\textbf{Experimental Results for PlaceLaundry Task.} The results demonstrate that HumanoidExo data is a viable substitute for teleoperated data, achieving comparable performance across both in-domain and out-of-domain scenarios.}
\label{tbl:PlaceLaundry}
\resizebox{0.45\textwidth}{!}{\begin{tabular}{cccc}
\toprule
 \makecell{Teleoperated \\ Data} & \makecell{HumanoidExo \\ Data}  & \makecell{Seen Cloth \\ Success Rate} & \makecell{Unseen Cloth\\ Success Rate}\\
\midrule

200 & 0 & 80\% & 80\%\\
5 & 195 & 80\% & 75\%\\
5 & 0 & 5\% & 5\%\\
\bottomrule
\end{tabular}}
\end{table}

\section{Conclusions}
In this work, we addressed the critical data bottleneck that hinders the development of capable, general-purpose humanoid robots. While existing methods like simulation, human videos, and direct teleoperation have advanced the field, they suffer from significant limitations in scalability, cost, and embodiment mismatch. We introduced HumanoidExo, a lightweight, wearable exoskeleton system designed to provide a practical and effective solution for scalable, whole-body data collection. Our experiments confirm that this approach is highly effective. We've shown that data from HumanoidExo enables policies to generalize to new environments, achieve remarkable data efficiency by learning complex skills from as few as five real-robot demonstrations, and even acquire entirely new skills like walking without any prior robot data. These results validate our system as a powerful paradigm for generating large-scale, high-quality humanoid datasets

{
    \small
    \bibliographystyle{ieeenat_fullname}
    \bibliography{main}
}


\end{document}